\pdfoutput=1

\documentclass[12pt]{article}

\usepackage[]{ACL2023}

\usepackage{times}
\usepackage{latexsym}
\usepackage{graphicx} 

\usepackage[T1]{fontenc}

\usepackage[utf8]{inputenc}

\usepackage{microtype}

\usepackage{inconsolata}
\usepackage{natbib}

\title{From Dialogue to Diagram: Task and Relationship Extraction from Natural Language for Accelerated Business Process Prototyping}

\author{Sara Qayyum 
   \And
  Muhammad Moiz Asghar \And
  Muhammad Fouzan Yaseen }

\begin{document}
{\makeatletter\acl@finalcopytrue
  \maketitle
}\begin{abstract}
The automatic transformation of verbose, natural language descriptions into structured process models remains a challenge of significant complexity - This paper introduces a contemporary solution, where central to our approach, is the use of dependency parsing and Named Entity Recognition (NER) for extracting key elements from textual descriptions. Additionally, we utilize Subject-Verb-Object (SVO) constructs for identifying action relationships and integrate semantic analysis tools, including WordNet, for enriched contextual understanding. A novel aspect of our system is the application of neural coreference resolution, integrated with the SpaCy framework, enhancing the precision of entity linkage and anaphoric references. Furthermore, the system adeptly handles data transformation and visualization, converting extracted information into BPMN (Business Process Model and Notation) diagrams. This methodology not only streamlines the process of capturing and representing business workflows but also significantly reduces the manual effort and potential for error inherent in traditional modeling approaches.
\end{abstract}

\section{Introduction}
In the dynamic domain of business process management (BPM), the task of translating detailed process descriptions from their natural, often verbose, linguistic form into structured, formal models stands as a critical yet arduous endeavor. This task, which has traditionally been dependent on manual interpretation and modeling by experts, is riddled with inherent challenges. Among these, the most prominent are the significant consumption of time and resources, the introduction of subjectivity and human bias from the varying resources carrying out a singular process, and the consequent high propensity for errors. Such challenges are compounded when dealing with complex, multifaceted business processes that are often described in a language rich with nuances and implicit meanings.

The emergence and continual advancement of Natural Language Processing (NLP) technologies herald a transformative shift in this landscape. Their application in the field of BPM is particularly significant, as it holds the promise of solving the varying challenges of converting unstructured text into structured process models, and with our efforts, we intend to create a holistic solution for automation. This automation is more than a mere translation of words; it involves an understanding of linguistic constructs, the identification and extraction of key process elements such as tasks, participants, decision points, and the contextual interpretation of these elements to ensure coherence in the resulting models.

Moreover, the ability of NLP to process and analyze language at scale can address the inefficiencies associated with manual modeling. By leveraging techniques such as semantic analysis, syntactic parsing, entity recognition, and co-reference resolution, NLP can sift through complex process narratives, distill essential information, and represent it in a standardized format like BPMN (Business Process Model and Notation). This capability not only enhances the accuracy and consistency of process models but also significantly expedites the process modeling exercise, thereby enabling organizations to rapidly document, analyze, and optimize their business processes.

In essence, the integration of NLP into BPM represents a confluence of linguistic intelligence and process management acumen, offering a pathway to more efficient, accurate, and scalable business process documentation. This intersection has the potential to fundamentally alter how organizations approach process modeling and optimization, marking a pivotal advancement in the field of business process management.

\section{Problem Statement}

To succinctly restate, at the heart of this research lies the ambitious endeavor of automating the transformation of unstructured, natural language descriptions of business processes into structured BPMN diagrams. This transformation is not merely a linear translation but a complex re-engineering of text into a formal, diagrammatic representation that captures the intricacies and nuances of business processes. Despite the remarkable advancements in NLP, the application of these technologies in the realm of business process modeling poses distinct and multifaceted challenges.

A primary challenge is the accurate identification and categorization of various entities and actions embedded in the text. Business process descriptions in natural language are typically riddled with specialized terminology, implicit assumptions, and complex constructs that vary significantly across different organizations and industries. Extracting meaningful and relevant information from such varied and intricate narratives demands an NLP system that goes beyond basic text parsing to understand the context and underlying semantics.

Moreover, understanding the relationships between extracted entities is critical. In business process narratives, entities do not exist in isolation; their interactions and interdependencies form the backbone of the process. The NLP system must, therefore, be capable of discerning these relationships, often implied or intricately expressed, and accurately reflect them in the BPMN diagrams.

Related challenges, as a result of the nuances of the English language, revolve around concepts like, 'Alias Detection', where diverse linguistic expressions often obscurely reference the same participant or entity. This issue necessitates an in-depth linguistic analysis to accurately identify participants amid numerous potential aliases. The second challenge pertains to the idea of 'Anaphora Resolution', which involves clarifying pronouns and other references in the text to mitigate interpretational ambiguity and ensure precision in the subsequent process mapping. 

Another significant challenge is the effective mapping of identified entities and relationships into a standardized business process framework. BPMN, being a globally recognized and widely used standard, has specific notations and conventions that must be adhered to. The translation from text to BPMN involves not only placing entities and actions into appropriate BPMN constructs but also ensuring that the flow and logic of the process are preserved. An envisioned solution would follow a pipeline similar to Figure 1.

\begin{figure}[ht]
  \centering
  \includegraphics[width=1\linewidth]{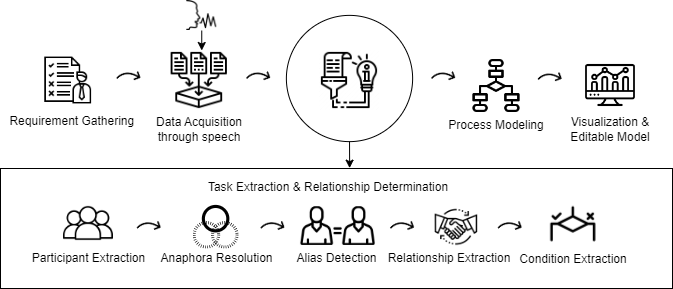}
  \caption{Proposed solution for task and relationship extraction from natural language}
  \label{fig:example}
\end{figure}

Lastly, and perhaps most critically, is the challenge of maintaining the semantic integrity of the original text during this transformation. The effectiveness of a BPMN diagram hinges on its fidelity to the original process description. Any loss of meaning or context in the translation can lead to inaccuracies in the diagram, potentially resulting in misinterpretation and misapplication of the process. Thus, the NLP system must be sensitive to the subtleties of language and context, ensuring that the BPMN diagrams it produces are not only structurally accurate but also semantically faithful to the source text. 

\section{Objectives and NLP Techniques in solving BPM Problems}

The overarching objective of this research initiative is to pioneer the development of a system with solution-oriented NLP techniques, each playing a pivotal role in the system's functionality.

Foremost among these techniques is dependency parsing, a critical tool for deciphering the syntactic structure of sentences. By discerning how words in a sentence relate to one another, dependency parsing enables the system to extract the core components of business processes, such as tasks, participants, and temporal sequences, with remarkable precision.

Complementing this is the use of Named Entity Recognition (NER), an NLP technique adept at identifying and classifying key terms within text. In the context of BPM, NER is invaluable for pinpointing specific business entities, actions, and resources, thereby laying the groundwork for their accurate representation in BPMN diagrams.

Another cornerstone of this system is neural coreference resolution. This advanced technique addresses the challenge of understanding pronouns and other referring expressions within the text, linking them back to the appropriate entities. The integration of neural coreference resolution ensures that the system can maintain narrative coherence across a business process description, a crucial factor for the accurate mapping of processes, addressing our earlier concerns of aliases and anaphoras within text.

Additionally, the system leverages the semantic analysis capabilities of tools like WordNet. By providing a rich semantic network of English words, WordNet allows the system to delve beyond surface-level text interpretation, enabling it to grasp the nuanced meanings and relationships inherent in business process descriptions.

\section{Review of Existing Approaches in the Literature}

The recent advancements in the intersection of NLP and BPM have led to innovative methodologies for transforming textual descriptions into structured BPMN models. This integration is crucial for automating and streamlining the process of BPMN generation, as illustrated by the breadth of research in this domain.

Chen et al. (2021) explored event-centric NLP, laying the groundwork for understanding complex event structures in text, which is essential for accurate BPMN model extraction. Their work aligns with Guimin et al. (2021), who delved into relation extraction with type-aware map memories, emphasizing the importance of understanding word dependencies in NLP for BPMN.

Loper and Bird’s (2002) introduction of the NLTK toolkit offered foundational NLP tools and methodologies, facilitating the analysis and processing of natural language, a key step in BPMN model generation. Sinha and Paradkar (2010) utilized use cases to convert process specifications into BPMN, demonstrating a practical application of NLP in BPM.

Kennedy and Boguraev (1996) addressed the challenge of pronominal anaphora resolution, a critical aspect in understanding and maintaining the coherence of textual descriptions for BPMN conversion. On the other hand, Epure et al. (2015) contributed by automating process model discovery from textual methodologies, showcasing the potential of NLP in extracting meaningful process information.

Honkisz, Kluza, and Wiśniewski (2018) proposed a novel concept for generating BPMN models from natural language, emphasizing the need for effective syntactic analysis in NLP for BPMN generation. Similarly, Van der Aa et al. (2019, 2020) focused on extracting declarative process models from natural language and further advanced the field by integrating speech recognition, enhancing the accessibility and intuitiveness of BPMN model generation. We also see this in the research of De Marneffe and Manning (2008) contributed to the syntactic parsing techniques essential for understanding natural language, a key step in BPMN model generation. Santoro, Borges, and Pino (2008) and Ghose, Koliadis, and Chueng (2007) explored narrative structures and process discovery from textual artifacts, enriching the methodologies for BPMN model generation from diverse text sources.

In conclusion, these studies collectively represent a significant leap forward in BPMN model generation from text. By addressing the complexities of natural language and leveraging various NLP techniques, these research efforts have not only streamlined the process of BPMN generation but also enhanced the accuracy and efficiency of these models, thereby making significant contributions to the field of business process management.

\section{Methodology}
Our multifaceted approach is not simply about deploying NLP algorithms in isolation but rather about orchestrating a series of interconnected and sophisticated steps, each critically contributing to the ultimate goal of automating and refining business process modeling.

To begin, the constants module in this project serves as an essential cornerstone for the NLP-based processing, encapsulating a sophisticated understanding of language structure crucial for analyzing business process descriptions. It meticulously defines a comprehensive set of enumerations, each pivotal for different aspects of syntactic and semantic analysis within the NLP framework. For instance, the Descriptor class outlines key grammatical roles and dependencies fundamental to dependency parsing, a vital NLP technique that deciphers the syntactic hierarchy and relationships between words in a sentence. This parsing is crucial for extracting meaningful information like tasks, participants, and actions from complex textual descriptions. Similarly, the VerbType and KeywordType classes categorize verbs and keywords that are instrumental in identifying specific actions and decision points in business processes, such as message events and conditional flows. These categorizations are critical in parsing business logic from natural language, enabling the system to distinguish and prioritize elements that are essential for BPMN modeling. 

A crucial aspect of the methodology involves Subject-Verb-Object (SVO) extraction, which enhances the approach introduced by Krzysztof et al. (2018). Hopping off of their approach of identifying key components from the text, i.e. 'Activity', 'Condition', 'Who', and 'Terminated' as seen in Table 1 - we enhance this approach by delving deeper into the different roles and enhancing contextual understanding of those roles through WordNet within our system - we also introduced industry specific roles as part of a larger project to inculcate newer roles not found in the earlier addressed framekwork. Additionally, we swapped the static condition extraction, the participant extraction methodology, and the general model creation through bpmn-python for more efficient and accurate alternatives, and the inclusion of swimlanes, not previously found in explored literature.

\begin{table*}[ht]
\centering
\small
\begin{tabular}{|c|p{4cm}|p{2.3cm}|p{3cm}|c|}
\hline
Order & Activity & Condition & Who & Terminated \\
\hline
0 & start & & & \\
1 & Follow Textbook Process & & Affairs Department & \\
2 & Inform Affairs Department & & Production Manager & \\
3a1 & Close Request &  Affairs Director rejects request &  Affairs Director & \\
3b1 & Document Required Knowledge &  Affairs Director approves request &  Affairs Director & \\
4 & Send Requirement & & Confidential Secretary & \\
5 & Check Status & & Affairs Director & \\
6 & & & & yes \\
\hline
\end{tabular}
\caption{Krzysztof et al. (2018)'s approach to creating a CSV layout for bpmn-python}
\end{table*}

These method employs advanced syntactic analysis techniques to identify SVO constructs within the text. The SVO extraction method is based on the concept that tasks performed by an entity can be effectively summarized in a Subject-Verb-Object format, comprising three key components:

\begin{enumerate}
  \item \textbf{Subject}: Represents the entity or actor initiating the action or task.
  \item \textbf{Verb}: Signifies the action or operation being performed.
  \item \textbf{Object}: Denotes the entity or object upon which the action is executed or the target of the task.
\end{enumerate}

SVO constructs are essential for representing the tasks and activities within business processes. They play a fundamental role in understanding the dynamics and interactions depicted in process narratives. By systematically extracting and analyzing SVO constructs, the methodology provides a structured approach to dissecting business process descriptions, facilitating more effective modeling, analysis, and optimization. This approach enhances the understanding of how various entities interact within the business context, ultimately improving process management and resource allocation.

More critically, the participant extractor module plays a critical role in meticulously identifying and extracting participant information from business process descriptions. This module employs targeted NLP techniques to recognize and categorize various entities, including users, systems, or roles mentioned in the text. Accurate categorization within the context of the business process is essential for constructing comprehensive process models that reflect all stakeholders accurately. However, in the process of identifying participants, the presence of pronouns or aliases referring to entities can hinder accurate identification of SVO constructs. To address this challenge, the neural coreference resolution technique was integrated into the system. The neural coreference resolution process involves identifying and replacing all aliases and pronouns that refer to a specific entity within the natural language text with the neuralcoref library. This ensures that the text is consistently represented in terms of the identified entities. Subsequently, this information serves as the basis for deriving roles and responsibilities within the business process, and aided in the creation of lanes within the pool of the BPMN diagram, contributing to the overall effectiveness of the participant extraction module.

The conversion module in conjunction with this handles the challenge posed by the heterogeneity and variability of data formats and structures inherent in natural language sources. It meticulously standardizes and prepares the textual data from these diverse formats, ensuring that it is in a suitable and uniform structure for detailed and intricate NLP analysis. This pre-processing step is pivotal; By standardizing the format and structure of the input text, this module ensures that the text is optimally primed for accurate interpretation and in-depth analysis by the subsequent NLP algorithms. We address the complex challenge of representing the fluid and often ambiguous constructs of human language in a structured, machine-processable format by creating clearly defined data models and structures for entities such as processes, tasks, or actors, this module translates the intricacies and subtleties of human language into a structured, organized format that can be effectively processed and analyzed by machines.

In our concluding note, we created a module to manually create the xml for the model generation as no current tools create a holistic BPMN with discernible roles, and while peripheral to the core NLP techniques, is vital in the context of data integration and handling. It focuses on the graphical representation of the processed data, including the creation of BPMN diagrams and other forms of visualizations. This module is critical for translating the structured process data, which has been meticulously analyzed and modeled through the application of various NLP techniques, into visual formats that are easily understandable and accessible to users. The ability to present complex process models and data in a visually intelligible and appealing format is essential for the practical application and usability of the insights gained from the NLP analysis. It enables users, who may not be experts in NLP or BPMN, to understand, interpret, and utilize the results of the process modeling exercise effectively.

In conclusion, the methodology adopted in this project exemplifies a deep and sophisticated understanding of both NLP and its practical applications in the realm of business process management. From the initial preparation and standardization of data to the intricate linguistic analysis, entity extraction, and the final visual representation, each step in the methodology is carefully designed and executed to contribute to the seamless translation of natural language into structured, actionable business process models. This approach not only demonstrates technical proficiency and innovation but also reflects a keen awareness of the practical needs and challenges in the field of business process management. 

\section{Results}
The application of the proposed methodology in this project successfully automated several critical aspects of transforming natural language process descriptions into structured BPMN diagrams. The results were notable in the areas of participant extraction, relationship extraction, alias detection, anaphora resolution, and condition/decision extraction, as a holistic solution. Each of these components played a pivotal role in accurately and efficiently modeling business processes. The creation of the model is seen in Figure 2.

\begin{figure}[ht]
  \centering
  \includegraphics[width=1\linewidth]{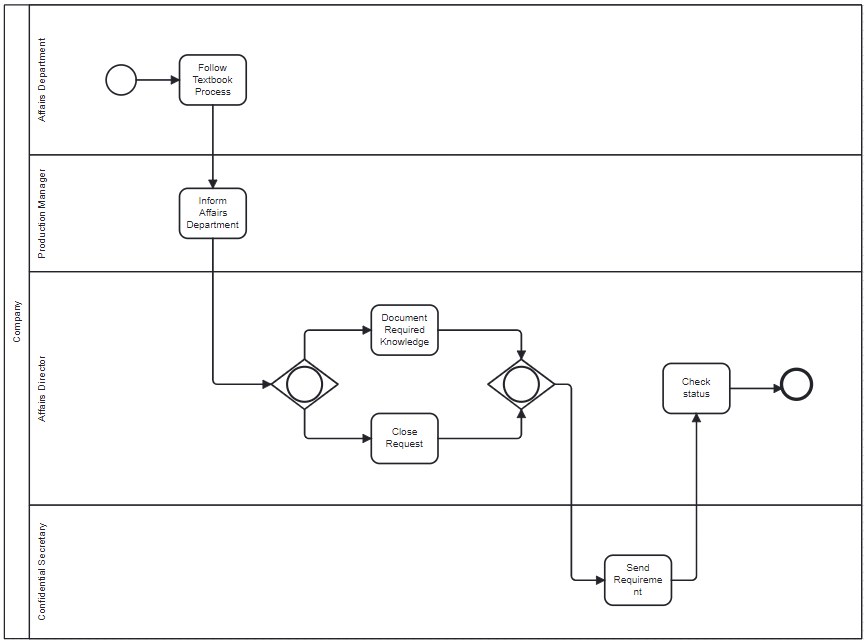}
  \caption{Generated BPMN against Table 1}
  \label{fig:example}
\end{figure}

The results of this project demonstrate the substantial potential of NLP techniques in automating and enhancing the process of business process modeling. The identification of tasks, participants, relationships, and decision points, coupled with effective alias and anaphora resolution, underscores the system's capability to transform natural language descriptions into comprehensive and accurate BPMN diagrams. These findings make a significant contribution to the field of BPM, particularly by introducing a role-based perspective in BPMN modeling. This novel approach, absent in previous literature, holds great significance within a business context as it facilitates a comprehensive understanding of how tasks align with the resources responsible for their execution, fostering improved cooperation and resource utilization.

\section{Discussion}
\subsection{Interpretation of Results}
The results obtained from this project provide a compelling insight into the efficacy of NLP techniques in the domain of business process modeling. The successful implementation of task identification, participant extraction, relationship extraction, alias detection, anaphora resolution, and condition/decision extraction signifies a major stride in addressing the core research question: How can NLP techniques be effectively utilized to transform unstructured natural language descriptions into structured BPMN diagrams?

The identification of tasks and participants, and the extraction of their interrelationships, directly contribute to creating comprehensive BPMN models that accurately mirror real-world business processes. This is a significant advancement, considering the complexity and nuances involved in interpreting unstructured text. The ability to detect aliases and resolve anaphoras further enhances the coherence and consistency of the BPMN diagrams, ensuring that they are not only accurate but also understandable and usable for business analysts.

Moreover, the extraction of conditions and decision points from the process descriptions is particularly noteworthy. This aspect of the system goes beyond mere structural modeling, delving into the logical and decision-making elements of business processes. It demonstrates the system's capability to handle not just the 'what' and 'who' of processes but also the 'how' and 'why,' which are often more challenging to model.

\subsection{Comparison with Existing Literature}
When compared with existing literature and methodologies in the field, the results of this project exhibit several distinct advantages. Traditional approaches to BPMN modeling often rely heavily on manual interpretation and conversion of process descriptions, which can be time-consuming and prone to errors. In contrast, the automated approach developed in this project significantly reduces the time and effort required for process modeling, while also improving accuracy and consistency.

Previous studies and tools in the domain, such as those discussed in the literature review, have demonstrated various degrees of success in automating parts of the business process modeling task. However, many of these approaches have limitations, particularly in handling complex linguistic structures, detecting nuanced relationships, and maintaining coherence in the models, and perhaps, more critically - poor identification of participants in lieu of creating lanes within the subsequent BPMN model. The comprehensive approach of this project, which integrates multiple advanced NLP techniques, addresses these limitations effectively. It not only automates the extraction of basic process elements but also successfully interprets and models complex linguistic and logical constructs present in natural language descriptions.

\section{Limitations}
While the project demonstrates significant advancements in employing Natural Language Processing (NLP) techniques for business process modeling, it is important to acknowledge certain limitations inherent in the approach.

\subsection{Dependence on Language Quality}
The system's accuracy and effectiveness heavily depend on the quality of the natural language input. In cases where process descriptions are poorly structured, ambiguous, or contain a high degree of colloquial language, the system may face challenges in accurately extracting process elements. This limitation highlights the need for well-articulated and clearly structured input, which may not always be feasible in real-world business scenarios.

\subsection{Handling of Complex and Nested Structures}
While the system performs well with standard and moderately complex process descriptions, it may struggle with highly complex or deeply nested linguistic structures. Processes involving multiple layers of decision-making or intricate participant interactions may not be fully captured by the current NLP techniques implemented. This limitation points to the need for further development and refinement of the algorithms to handle such complexities more effectively.

\subsection{Limitations of the SVO approach}
In the context of task identification, while SVO constructs provide a structured framework for representing actions within business processes, it is important to acknowledge that SVO alone may not always be the best approach. Tasks within complex business processes can exhibit intricate dependencies and nuanced interactions that may not be fully encapsulated by a simplistic SVO format. SVO constructs tend to oversimplify the statement creating unnatural representations of the task at hand. In scenarios where tasks involve conditional actions, parallel processing, or intricate decision trees, relying solely on SVO constructs may lead to an incomplete and inaccurate representation of the actual process dynamics. Therefore, a more comprehensive approach will produce more natural and accurate tasks for the BPMN.

\subsection{Generalizability Across Languages}
The current implementation primarily focuses on English language process descriptions. The system’s applicability and effectiveness in other languages have not been fully explored. This limitation is significant in a global context where business processes often span multiple languages and cultural contexts.

\section{Conclusion}
This project represents a significant advancement in the application of NLP techniques for transforming natural language process descriptions into structured BPMN diagrams. The methodology developed and the results obtained underscore the potential of NLP in automating and refining business process modeling, a task traditionally characterized by its manual intensity and complexity.

The system's ability to accurately identify tasks, extract participants, delineate relationships, detect aliases, resolve anaphora, and extract conditions and decision points from natural language text is a testament to the power of NLP. It not only enhances the efficiency and accuracy of BPMN modeling but also offers a deeper, more coherent interpretation of process descriptions than many existing methods.

However, as acknowledged in the limitations section, the system does have constraints, including its dependence on the quality of the input language, contextual understanding, scalability, and language generalizability. These limitations offer valuable insights and directions for future research. Addressing these challenges could lead to even more sophisticated NLP systems capable of handling a wider variety of business process modeling scenarios, including those involving real-time processing, complex decision-making structures, and multilingual environments.

In conclusion, this project paves the way for further advancements in the field of BPM, opening new avenues for research and development. The integration of advanced NLP techniques in business process modeling has the potential to significantly transform how organizations document, analyze, and optimize their processes. As NLP technology continues to evolve, its integration into BPM tools is expected to become more refined, making these tools even more powerful and essential in the world of business process management.

\nocite{*}
\bibliographystyle{plainnat} 
\bibliography{FromDialogueToDiagram} 

\end{document}